\documentclass{article}
\PassOptionsToPackage{numbers, compress}{natbib}
\usepackage[preprint]{neurips_2021}
\usepackage[utf8]{inputenc} 
\usepackage[T1]{fontenc}    
\usepackage{hyperref}       
\hypersetup{
    colorlinks=true,
    linkcolor=blue,
    filecolor=magenta,      
    urlcolor=cyan
    }
\usepackage{url}            
\usepackage{booktabs}       
\usepackage{amsfonts}       
\usepackage{nicefrac}       
\usepackage{microtype}      
\usepackage{xcolor}         
\usepackage{graphicx}
\usepackage{mathtools}
\usepackage{amssymb}
\usepackage{amsmath}
\usepackage[ruled,vlined]{algorithm2e}
\usepackage{multicol}
\usepackage{multirow}
\usepackage{subfig}

\title{Causal Discovery in Knowledge Graphs by Exploiting Asymmetric Properties of Non-Gaussian Distributions}

\author{%
  Rohan~Giriraj, Sinnu~Susan~Thomas\\
  Department of Computer Science and Engineering\\
  Digital University Kerala (IIITMK)\\
  India 695317 \\
  \texttt{rohan.mi19, sinnu.thomas@iiitmk.ac.in} \\
}

\begin{document}

\maketitle

\begin{abstract}

In recent years, causal modelling has been used widely to improve generalization and to provide interpretability in machine learning models. To determine cause-effect relationships in the absence of a randomized trial, we can model causal systems with counterfactuals and interventions given enough domain knowledge. However, there are several cases where domain knowledge is almost absent and the only recourse is using a statistical method to estimate causal relationships. While there have been several works done in estimating causal relationships in unstructured data, we are yet to find a well-defined framework for estimating causal relationships in Knowledge Graphs (KG). It is commonly used to provide a semantic framework for data with complex inter-domain relationships. In this work, we define a hybrid approach that allows us to discover cause-effect relationships in KG. The proposed approach is based around the finding of the instantaneous causal structure of a non-experimental matrix using a non-Gaussian model, i.e; finding the causal ordering of the variables in a non-Gaussian setting. The non-experimental matrix is a low-dimensional tensor projection obtained by decomposing the adjacency tensor of a KG. We use two different pre-existing algorithms, one for the causal discovery and the other for decomposing the KG and combining them to get the causal structure in a KG. 
\end{abstract}

\section{Introduction}
\label{introduction}

Inclusion of causality in fields such as machine learning has shown improved results when it comes to problems such as generalizations and countering adversarial attacks \cite{yang2019causal}. 
Causality has been applied to the different modalities of data required in machine learning, be it images \cite{zhu2020cookgan}, text \cite{keith-etal-2020-text} or numerical data \cite{spirtes2000causation}. However, of these modalities, much work is not done in terms of causal discovery in KG \textemdash semantically rich method for storing and retrieving relational data. They describe entities based on the relationships they hold with other entities. These entities and relationships combined are represented in the form of a tuple $(s,p,o)$ (subject, predicate, object) or $(h,r,t)$ (head, relation, tail), which is called a triple. Eq. \eqref{triple} shows the diagrammatic representation of a triple.
\begin{equation}
\label{triple}
subject \xrightarrow{predicate} object
\end{equation}
KG possesses statistical properties that show how some triples affect other triples i.e; the existence of some triples affects the existence of certain other triples. This statistical dependence between the triples of a KG leads us to postulate the existence of a cause-effect relationship that governs how the triples affect each other. Our assumption is based on the Common Cause Principle \cite{hofer1999reichenbach}, which states that when two variables $X$ and $Y$ are statistically dependent, there exists a third variable $Z$ that influences both $X$ and $Y$. This third variable is called a ``confounder" and in most cases, remains ambiguous. In the proposed method, we assume that there are no latent confounders, and only the observed data is taken into consideration. The assumption is necessary for the causal discovery algorithm, Linear Non-Gaussian Acyclic Model (LiNGAM) \cite{JMLR:v7:shimizu06a} that discovers the causal order using Independent Component Analysis (ICA) \cite{Hyvaerinen2000}  under non-Gaussian assumption on observational data. LiNGAM exploits the asymmetry of non-Gaussian data in higher order statistics, to determine the causal direction among the data points.

The implications of causal discovery in the context of KG are huge, allowing better path discovery for explainable reasoning and querying. 
In this work, we define a hybrid theoretical approach to attempt solving the problem of causal discovery in KG. The contributions of this work are threefold. At first, we use TuckER \cite{Balazevic2019}, a method for embedding a KG after decomposing its adjacency tensors into a core tensor and constituent matrices. Secondly, we take the decomposed tensor, and we project them into a matrix, \textbf{Q}. Finally, the projected matrix is then passed to the DirectLiNGAM \cite{Shimizu2011} algorithm that finds the causal ordering from the given data matrix.

The remainder of this paper is organized as follows. Section \ref{lit_review} contains a brief literature review and underscores the proposed contributions. Section \ref{methodology} introduces the causal discovery in KG. Section \ref{results} features numerical experiments and Section \ref{limit} highlights the limitations of the proposed work. Finally, Section \ref{conclusions} concludes the work.

\section{Literature Review} 
\label{lit_review}
The proposed work tries to define a framework that can infer causal relationships from KG \textemdash a multi-relational graph composed of entities and relationships which are regarded as nodes and different types of edges respectively \cite{wang2017knowledge}. They are known to possess certain statistical properties such as transitivity and type constraints \cite{Ji2021tnnls} and also some softer statistical patterns can be seen in larger KG. YAGO \cite{yago}, DBPedia \cite{dbpedia}, Freebase \cite{fb15k}, WordNet \cite{fb15k} etc. have given researches access to large KGs which allowed them to experiment and discover more statistical patterns within them. We need to represent the KG in such a way that it becomes easier for us to capture the statistical relationships present in the KG. Most models represent triples via latent entities. Such representations can be broadly classified into three main categories: point-wise space, complex space, Gaussian space and Manifold space \cite{Ji2021tnnls}.

Of these, we concentrate on pointwise space representation of the KG that includes vector, matrix, and tensor space in this work. TransE \cite{fb15k} represents the translational representation of entities in $d$-dimensions $h+r \approx t$. TransR \cite{TransR} was introduced to combat the problem of representing both the entities and relationships in single space and defines a matrix $M_r \in \mathbb{R}^{k  \times d}$, where $k$ is the entity embedding space and $d$ is the relation space. TransH \cite{wang2014knowledge} is another example where the translational model is extended to work with hyperplanes. Other methods such as HolE \cite{HolE} which is based on semantic matching, uses a plain vector space for its representation.  Encoding models are just better versions of the point-wise space models, where simple models such as bilinear models \cite{NIPS2012_0a1bf96b} achieve state-of-the-art performance compared to the existing point-wise space models such as TransE, RESCAL \cite{rescal}, DistMult \cite{yang2015embedding}, and ComplEx \cite{ComplEx}. Of these models, we are most interested in RESCAL, which explains triples as pairwise interactions of latent features. RESCAL can be further extended to work with higher dimensional data to handle entities properly \cite{yago_factorizing}. A better method for encoding relationships is TuckER that was introduced to decompose the KG using three-way tucker decomposition \cite{Balazevic2019} to a core tensor and constituent matrices of entities and relationships.

We choose TuckER over other methods since it allows full-expressivity and other encoding models such as RESCAL that can be considered as a special case of TuckER. It also takes the asymmetry of the KG into consideration since the proposed causal inference method is dependent on that. While there are several other encoding methods such as the neural network based NTN \cite{NTN} and NAM \cite{liu2016probabilistic}, we follow Occam's Razor and a relatively simpler linear model of TuckER architecture for the experiments. Causal discovery algorithms can be classified into three categories. First, constraint based (CB) algorithms learn a set of causal graphs that satisfy the conditional independence in the data. Statistical tests can be used to verify if a candidate graph is \textit{faithful} \cite{spirtes2000causation}. A popular example of this is the Peter-Clark algorithm \cite{spirtes2000causation}. Second, score based (SB) algorithms check for the goodness of the fit tests instead of testing for conditional independence. It uses a scoring function, Bayesian Information Criterion (BIC) \cite{schwarz1978estimating} and maximizes the criterion. There are certain hybrid algorithms that combine both SB and CB methods, for achieving a better result. Third, structural causal model (SCM) based algorithms in which a variable can be written as a function of the directed causes and some noise term.

We consider the \textit{Pearlean} \cite{Pearl2009} approach for causal inferences, in the context of SCMs \cite{Pearl2009} in this work. SCM provides a comprehensive theory of causality. Eq.\eqref{scm} shows how SCM can be represented as a function, also called as Functional Causal Model (FCM)
\begin{equation}
\label{scm}
    X_i = f_i(\textbf{PA}_i, U_i) 
\end{equation}
where $f_i$ is a deterministic function, $\textbf{PA}_i$ is the parent(s) of the $X_ith$ observable random variable. $U_i$ is an exogenous variable that is assumed to be jointly independent.
SCMs consist of two components: the causal graph and the structural equation \cite{bollen1998interactions}. In causal graphs, each node denotes a random variable and each directed edge from $X$ to $Y$ denotes the causal influence of $X$ on $Y$. Causal graphs are usually assumed to fulfill the Markov Property such that the implied joint probability factorizes into a “disentangled representation” following recursive decomposition. In the case of causal graphs, there can be cases where multiple graphs can satisfy the conditional independencies. To identify the scenarios where the true graph is identifiable, for which the most common example is a linear system based on non-Gaussian errors \cite{JMLR:v7:shimizu06a}.

Non-Gaussianity is asymmetric when we take higher order statistics into account \cite{Dodge2001,Dodge2009}. We use an algorithm that specifically exploits this asymmetric property to find the causal order, LiNGAM. The task of learning in LiNGAM comes down to estimating the lower triangular matrix that shows the causal order $k(j)$ where no variable precedes its cause,  similar to topologically sorting Directed Acyclic Graph (DAG) \cite{pang2015topological}.  LiNGAM uses ICA  to decompose the data matrix $\textbf{X}  = \textbf{BS}$, where $B$ is the mixing matrix and $S$ is the source matrix. The matrix $\textbf{B}$ is then used to compute $\textbf{W} = \textbf{B}^{-1}$, which is the unmixing matrix. $\textbf{W}$ can then be used to find the lower-triangular matrix. ICALiNGAM gets stuck at the local optima rather than the global ones. To counter this, DirectLiNGAM was introduced, which guaranteed convergence. Although its approach is more or less similar to ICALiNGAM, it uses Kernelized ICA \cite{bach2002kernel} by kernelizing the canonical correlation \cite{akaho2006kernel} of the variables.

The proposed method is heavily inspired by the  multi-dimensional causal discovery \cite{schaechtle2013multi} that introduces a tensor decomposition based around tucker decomposition and HOSVD \cite{wang2017tensor} for discovering causal structures in temporal data.
Another work that explicitly discussed the problem of causal inference in KG \cite{semex_2019_4}  which is based on the idea of pruning KG and having a probabilistic relational model learn the causal structures within the pruned graph. 
However, pruning the graph requires explicit domain knowledge and insight on how the KG is designed.  This is where the proposed method is different, as we don't require any domain data for causal discovery. This makes the proposed approach more robust and generalizable for multiple datasets and domains. Other methods are very interactive, requiring human input occasionally, but the proposed method can find the causal relationships between the embeddings of the KG.

\section{Methodology} \label{methodology}
\subsection{Causal Discovery Framework}

We formulate the causal discovery framework for the given variables to discover their causal order. We use a non-Gaussian version of Structural Equation Model (SEM) and Bayesian Network (BN) called LiNGAM which approximates a causal order $k(i)$ for a set of observed variables modeled as a DAG. LiNGAM is based around some key assumptions:
\begin{itemize}
\item The set of observed variables ${x_1,...,x_n}$ can be arranged in a causal order, such that no later variable causes any earlier variable. A causal order is denoted as $k(i)$.

\item Each observable $x_i$ can be designed as a linear combination of its earlier variables. 
\begin{equation}
\label{lingam}
    x_i = \sum_{k(j)>k(i)} b_{ij}x_j + e_i
\end{equation}
Here, $b_{ij}$ represents the connection strength of the variable $x_i$ and $x_j$, $e_i$ represents a noise term. In the case of KG, we assume the relationships between the triples to be of linear nature.

\item The disturbances $e_i$ are continuous, independent random variables with non-Gaussian distributions. This is the same as the noise variable $U_i$ from Eq. \ref{scm}.

\end{itemize}

The above assumptions imply that there are no latent confounders involved in the proposed system, and this is called “causally sufficient”. In our specific case, it means that there are no unobserved triples in a KG. 

\subsection{Why Non-Gaussianity?}
According to Shimizu et al. \cite{JMLR:v7:shimizu06a}, algorithms based around second-order statistics are unable to discern the entire causal structure in most cases. A simple example would be the consideration of two variables $a$ and $b$ where they both are statistically dependent on each other. We know for a fact that in the case of a Gaussian distribution, $\rho_{ab}$ — the Pearson correlation coefficient, is symmetric which means that the direction of dependence, $a \rightarrow b$ or $b \rightarrow a$, is unidentifiable. However, in higher order statistics, the correlation coefficients exhibit properties of asymmetry \cite{Dodge2001,Dodge2009}.
The \textit{fourth standard moment}- kurtosis of a distribution is defined as 
\begin{equation}
    \kappa_x = \textbf{E}\Big[\Big(\frac{X-\mu_x}{\sigma_x}\Big)^4\Big] = \frac{\mu_4}{\sigma^4}
\end{equation}
where $\mu_4$ is the standardized central moment, \textbf{E} is the expectation and $\sigma$ is the standard deviation. 
Excess Kurtosis can be defined as $\kappa_x-3$. The fourth power of $\rho_{XY}$ can be written as the ratio of the excess kurtosis of response and predictor, where response is $Y$ and predictor is $X$, when $X \rightarrow Y$ is considered.
\begin{equation}
\rho_{xy}^4 = \frac{\kappa_y}{\kappa_x}
\end{equation}

$\rho_{xy}$ is bounded by the interval $[-1,1]$.
So when we consider a Gaussian distribution, the excess kurtosis will be $\kappa_x = 0, \kappa_y = 0$, resulting in an irrational value $\frac{0}{0}$. Therefore, we know that asymmetry exists in the case of non-Gaussian distributions when higher moments are taken into consideration.

Traditional LiNGAM models the problem of causal discovery in the form of ICA. ICA uses a set of inverse linear basis transformations to generate the constituent components of a mixed signal. ICA can be modeled as matrix multiplication $x(t) = \textbf{M}s(t)$, where $x(t)$ and $s(t)$ are the observed and source signals respectively and $\textbf{M}$ is the unknown mixing matrix.
Eq. \eqref{lingam} can be modeled in the same form as above, $x = \textbf{A}e$ where $\textbf{A} = (\textbf{I}-\textbf{B})^{-1}$ where $\textbf{B}$ is the mixing matrix in the ICA problem setting and $I$ is the identity matrix. This form, taken along with the non-Gaussian and independent components of \textbf{e}, is called the linear independent component analysis model.

However, there are problems with the ICA approach. The calculation of gradients, to minimize the diagonal elements of the unmixing matrix \textbf{W}, does not guarantee convergence of the solution. There is a risk of the algorithm getting stuck at the local optima. Secondly, the permutations performed to obtain the lower-triangular matrix are scale-invariant, leading to the finding of the wrong causal order. To mitigate these issues, we prefer DirectLiNGAM that guarantees convergence. Compared to the pre-existing methods this algorithm requires no algorithmic parameters and is guaranteed to converge to the right solution within the fixed number of steps, provided it follows the model strictly along with all the assumptions and the sample size is infinite.

In DirectLiNGAM, the input is a $p$-dimensional vector $x$ with variable subscripts $U$ and a data matrix $X$ of shape $p \times n$. Then two ordered lists, $K := \emptyset$ and $m := 1$ are initialized.
\\
Then until $p-1$ subscripts have been appended to $K$:
\begin{itemize}
\item Least square regression is done on variables $x_i$ and $x_j$, where $i \in U \setminus K (i\neq j)$ and the residual vectors $r^{(j)}$ and the residual data matrix $R^{(j)}$ from $X$.
Then a variable $x_m$ is computed by minimizing an independence measure.
\begin{equation}
x_m = \arg \underset{j \in U \setminus K}{\min} T_{kernel}(x_j;U\setminus K)
\end{equation}
$T_{kernel}$ is a kernelized measure of independence computed by calculating the mutual information between the variable $x_j$ and residual vector $r_i$

\begin{equation}
    T_{kernel}(x_j;U) = \sum_{i \in U, i \neq j} \widehat{MI}_{kernel}(x_j, r_i^{(j)})
\end{equation}
The kernel-based mutual information estimator can be written as:

\begin{equation}
\widehat{MI}_{kernel}(y_1,y_2) = -\frac{1}{2}\log\frac{\det \mathcal{K}_\tau}{\det \mathcal{D}_\tau}
\end{equation}

where $\tau$ is a small positive constant and $\mathcal{K}_\tau$ and $\mathcal{D}_\tau$ are just matrices whose blocks are $(\mathcal{K}_\tau)_{ij} = K_iK_j$ for $i\neq j$, and $(\mathcal{K}_\tau)_{ii} = (K_i + \frac{N\tau}{2}I)^2$ and $\mathcal{D}_\tau$ is just a block diagonal matrix with blocks $(K_i+\frac{N\tau}{2}I)^2$. $K_1$ and $K_2$ are just Gram matrices whose elements are RBF kernels of the sets of observations of $y_1$ and $y_2$ respectively (in the two variable case). 

\item $m$ is appended to $K$.
\item $x:=r^{(m)}$, $X := R^{(M)}$
\end{itemize}

Once $p-1$ subscripts have been appended to $K$ the remaining variable is appended to $K$ and a strict lower-triangular matrix $B$ is formed using the order in $K$, and the connection strengths $b_{ij}$ is estimated by a conventional regression method such as least squares or maximum likelihood based on covariance.

\subsection{KG Representation}
KG are usually represented in the form of triples. Triples consist of entities and relationships. Let the set of all entities be  $\mathcal{E} = \{e_1,...,e_{N_e}\}$,
and the set of all relationships be $\mathcal{R} = \{r_1,...,r_{N_r}\}$. The triples are usually represented as tuples $(e_i,r_k,e_j)$. Each possible triple can also be modeled as a binary random variable $y_{ijk} \in \{0,1\}$.

\[
y_{ijk} = 
    \begin{dcases}
        1, & \text{if the triple $(e_i,r_k,e_j)$ exists}\\
        0 & \text{otherwise} \\
    \end{dcases}
\]


This can also be represented as a third order adjacency tensor of shape \textbf{Y}$=\{0,1\}^{N_e \times N_e \times N_r}$. Estimation of the joint probability distribution of the observed triples can help us derive the entire KG. For our particular application, we assume that the KG is built in Closed World Assumption (CWA) \cite{minker1982indefinite}.
We know that some triples are statistically dependent on others, meaning that the existence of some triples influences the existence of other triples. To model this correlation, we need to assume that all triples $y_{ijk}$ are conditionally independent given the latent features associated with the subject, predicate, object and observed graph features and parameters. These score-based models predict the feasibility of a triple based on a scoring function $f(x_{ijk}, \theta)$, where the score denotes the confidence that a triple exists given the parameters $\theta$. In most cases, the triples are explained by the latent features, where latent features are the elements in a vectorized form of an entity/relationship in a triple. These latent features can come from different embedding algorithms that essentially convert text into vectorized form. For example: consider a system with only two entities then the latent features can be written in vectorized form as:
\begin{equation}
    e_1 =
    \begin{bmatrix}
    0.98\\
    0.
    \end{bmatrix},
    e_2 = 
    \begin{bmatrix}
    0.\\
    0.2
    \end{bmatrix}
\end{equation}

There are many more complicated methods for embedding entities, which we discussed in the Section \ref{lit_review}. These are language models that take in some text and output the resultant vector with a certain number of latent features.
The main intuition behind the latent feature model is the fact that the relationships between the different entities can be deduced by the interactions between their latent features.
In the proposed approach, we use these latent features to predict the causal order of the system.
There are several score-based methods such as RESCAL, DistMult, and ComplEx  but we prefer TuckER for the proposed approach. It is fully expressive, meaning that it is capable of capturing all the information present within the data. It allows us to represent all the other score-based, tensor factorization models as special cases of itself. It is a linear model, which is important because one of the major assumptions required for LiNGAM is linearity.

Tucker decomposition decomposes a tensor into its constituent matrices and a smaller core tensor. In a three mode scenario a tensor $\mathcal{X} \in \mathbb{R}^{I \times J \times K}$ outputs a core tensor 
$\mathcal{Z}$, and factor matrices \textbf{A}, \textbf{B} and \textbf{C}. Elements of the core tensor show the level of interaction between the different components.
\begin{equation}
    \mathcal{X} \approx \mathcal{Z} \times_1 \textbf{A} \times_2 \textbf{B} \times_3 \textbf{C}
\end{equation}

\begin{figure}%
    \centering
    \subfloat[\centering Core tensor of TuckER]{\label{core-tensor}{\includegraphics[width=5cm]{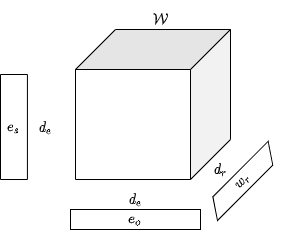} }}%
    \qquad
    \subfloat[\centering Visualization of tensor $\mathcal{Q}$]{\label{tensorQ}{\includegraphics[width=5cm]{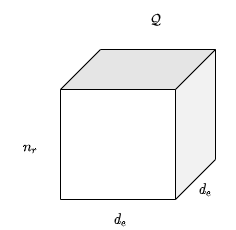} }}%
    \caption{\ref{core-tensor}: Visualization of the TuckER architecture, where $e_o, e_s \in \mathbb{R}^{d_e}$ and $w_r \in \mathbb{R}^{d_r}$.\\ 
    \ref{tensorQ}: Visualization of the tensor $\mathcal{Q} \in \mathbb{R}^{n_r \times d_e \times d_e}$}%
    \label{kg-rep}%
\end{figure}

We take the binary adjacency tensor representation of the KG and use Tucker decomposition to reduce it to the core tensor and factor matrices. We have an entity embedding matrix \textbf{E} which is equivalent for both subject and object entities, i.e; $\textbf{E} = \textbf{A} = \textbf{C} \in \mathbb{R}^{n_e \times d_e}$. The second factor matrix is the relation embedding matrix $\textbf{R} = \textbf{B} \in \mathbb{R}^{n_r \times d_r}$ where $n_e$ and $n_r$ are the number of entities and relations respectively and $d_e$, $d_r$ are the dimensions of the entity and relation matrices. These dimensions are the same as the latent features we mentioned above.
Finally, we have a core tensor $\mathcal{W} \in \mathbb{R}^{d_e \times d_r \times d_e}$, which is diagrammatically represented in Fig. \ref{core-tensor}. The scoring function of the TuckER is:
\begin{equation}
    \phi(e_s,r,e_o) = \mathcal{W} \times_1 e_s \times_2 w_r \times_3 e_o
\end{equation}
where $e_s,e_o \in \mathbb{R}^{d_e}$ are rows of the entity embedding matrix \textbf{E} denoting the subject and object entities of a triple. $w_r \in \mathbb{R}^{d_r}$ are the rows of relation embedding matrix \textbf{R} denoting the relations or the predicate values between the different entities. The result of the scoring function is then passed through the logistic sigmoid function so that the probability of a triple being true can be calculated. The final score obtained is $\sigma(\phi(e_s,r,e_o))$. We need to have a matrix that can be used to represent the decomposed form of a tensor apart from the core tensor and different factor matrices.
To achieve this, we introduce a projection tensor $\mathcal{Q} \in \mathbb{R}^{n_r \times d_e \times d_e}$ defined as
\begin{equation}
\label{q_tensor_eq}
    \mathcal{Q} = (\mathcal{W} \times_1 \textbf{E} \times_2 w_r) \textbf{E}.
\end{equation}
This tensor, as shown in Fig. \ref{tensorQ} captures the relationship between the dimensions of the entity embeddings $d_e$, since we are more interested in the latent variables that make up the entities. The resultant tensor $\mathcal{Q} \in \mathbb{R}^{n_r \times d_e \times d_e}$ is converted into a matrix $\textbf{Q} \in \mathbb{R}^{n_r \times (d_e\times d_e)}$ suitable for DirectLiNGAM.

\subsection{The Hybrid Algorithm}
We take the matrix \textbf{Q} and use that as a $p \times n$ dimensional data matrix as input for DirectLiNGAM. As a final check, we perform a kurtosis test and find that the matrix is non-Gaussian. DirectLiNGAM has a run time complexity of $\mathcal{O}(np^3M^2 + p^4M^3)$, so depending on the size of $p$, the runtime of the algorithm increases polynomially. For higher values of $p$, we reshaped the values of \textbf {Q} accordingly. The proposed approach gives the freedom to choose the number of relationships we can consider for the algorithm i.e; we can choose the number of rows of \textbf{R}, $w_r$ for some added flexibility as given in Algorithm \ref{hybrid_algorithm}.

\begin{algorithm}[H]
\label{hybrid_algorithm}
\SetAlgoLined
\KwResult{Causal order $k(i)$ }
\begin{itemize}
    \item Initialize the DirectLiNGAM algorithm with the right parameters.
    \item Initialize the triples of a KG as a third order tensor.
    \item Apply the TuckER model on the tensor to obtain the decomposed components.
    $\mathcal{Z},\textbf{A},\textbf{B},\textbf{C} \leftarrow$ TuckER(adjacency tensor)
    \item Choose the number of rows $w_r$ of relation matrix \textbf{R} we need for the execution.
    \item Get the projection tensor $\mathcal{Q} \leftarrow   (\mathcal{W} \times_1 \textbf{E} \times_2 w_r)\textbf{E}$.
    \item Matricize the tensor $\mathcal{Q}$ into the matrix $\textbf{Q}$.
    \item Causal order $k(i)$ $\leftarrow$ DirectLiNGAM(\textbf{Q})
    \item return $k(i)$
    
\end{itemize}
 \caption{Hybrid Algorithm}
\end{algorithm}

\section{Experimental Results} \label{results}
As the proposed approach is a highly specific application of causal discovery, there are no known benchmarks to check for causality in a KG that we can compare the proposed approach to. We find that it does give us a causal ordering of $d_e$, and an adjacency matrix, which we could then plot as a DAG. The code for the implementation is available  \href{https://www.github.com/rohangiriraj/CausalKG}{here}.

\subsection{Implementation}
\begin{itemize}
    \item \textbf{Datasets:} For the testing, we chose two datasets FB15K-237 \cite{fb15k-237}, and WN18-RR \cite{wn18rr} subsets of the FB15K \cite{fb15k} and WN18 \cite{fb15k} datasets respectively, but have their inverse relations removed. The removal of inverse relations helps us avoid cycles in the data. For our experimentation purposes, we chose a $w_r$ value of $100$ for FB15K-237 and $10$ for WN18-RR.
    Fig. \ref{inverse} shows an example of an inverse relation in a KG.  Note that the entities and the relationships are the same.
    \begin{figure}[h!]
        \centering
        \includegraphics[scale = 0.65]{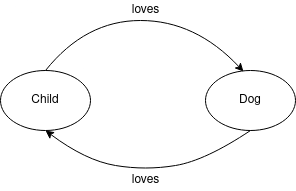}
        \caption{Inverse relations in a KG.}
        \label{inverse}
    \end{figure}

    \item \textbf{Libraries:} We use a combination of different open-source libraries to implement the algorithm. We use \texttt{numpy} \cite{harris2020array} for the tensor operations, \texttt{pykg2vec} \cite{pykg2vec} for overall embedding and tuning with TuckER, \texttt{lingam} \cite{JMLR:v7:shimizu06a} library for working with DirectLiNGAM and \texttt{graphviz} for visualizing the DAG from the causal order. 
    
    \item \textbf{Hardware:} For training of the TuckER architecture and the implementation of DirectLiNGAM, we use $9^{th}$ generation Intel Core i5 9300H processor, with 8 Gigabytes of RAM. This didn't particularly slow the experimentation down, as we considered smaller dimensions and smaller subsets of the larger KG datasets.
\end{itemize}

\subsection{Results}
\begin{table}[h!]
\resizebox{\textwidth}{!}{%
\begin{tabular}{|c|l|c|c|c|c|c|c|}
\hline
\multirow{2}{*}{Dataset}   & \multicolumn{1}{c|}{\multirow{2}{*}{$n_{dim}$}} & \multicolumn{2}{c|}{Convergence} & \multicolumn{2}{c|}{Execution Time}                                & \multicolumn{2}{c|}{Mean $p$-values}                               \\ \cline{3-8} 
                           & \multicolumn{1}{c|}{}                         & ICALiNGAM     & DirectLiNGAM     & \multicolumn{1}{l|}{ICALiNGAM} & \multicolumn{1}{l|}{DirectLiNGAM} & \multicolumn{1}{l|}{ICALiNGAM} & \multicolumn{1}{l|}{DirectLiNGAM} \\ \hline
\multirow{3}{*}{FB15K-237} & $ d_e = 5$                                    & No            & Yes              & 1.417                          & 3.957                             & 0.158                          & 0.200                             \\ \cline{2-8} 
                           & $d_e = 7$                                     & No            & Yes              & 8.821                          & 28.27                             & 0.162                          & 0.123                             \\ \cline{2-8} 
                           & $d_e = 10$                                    & Yes           & Yes              & 23.85                          & 240.4                             & 0.150                          & 0.183                             \\ \hline
\multirow{3}{*}{WN18-RR}   & $d_e = 5$                                     & No            & Yes              & N/A                            & 3.330                             & N/A                            & 0.335                             \\ \cline{2-8} 
                           & $d_e = 7$                                     & No            & Yes              & N/A                            & 24.08                             & N/A                            & 0.260                             \\ \cline{2-8} 
                           & $d_e = 10$                                    & No            & Yes              & N/A                            & 202.5                             & N/A                            & 0.153                             \\ \hline
\end{tabular}%
}

\caption{Quantitative Analysis with different parameters.}
\label{results}
\end{table}

Table \ref{results} shows four primary columns with different parameters specified in each.
\begin{itemize}
    \item \textbf{$n_{dim}$:} This specifies the dimensions of the embedding, i.e; $d_e$, $d_r$. For the sake of simplicity, we are considering a case where $d_e = d_r$. We tested with $5$, $7$ and $10$ dimensions. Since the dimensions grow quadratically, it makes sense to experiment with smaller values of $d_e$. The significance of smaller dimensions is downplayed by the non-convergence of the algorithms due to lack of data.
    
    \item \textbf{Convergence: } This shows whether the algorithm in question converges to provide a solution. We observe that ICALiNGAM fails to converge in most cases, especially in the case of smaller datasets like WN18-RR. This further proves the significance of a method like DirectLiNGAM, which is guaranteed to converge. In all the trials, we found that the ICA algorithms like FastICA were unable to converge despite early stopping. Since it converges at the local minima, there are chances for the causal order of ICALiNGAM to be wrong.
    
    \item \textbf{Execution time:} This specifies the execution time taken by both ICALiNGAM/LiNGAM and DirectLiNGAM to compute the causal order $k(j)$. Here we observe that in all cases, the execution time of ICALiNGAM is far lesser than that of DirectLiNGAM. This is because there is a trade-off between execution speed and convergence. While LiNGAM is faster, it failed to converge.
    If given enough computing resources and time, DirectLiNGAM is the better choice. 
    \item \textbf{Mean $p$-value:} This is the mean of the $p$-values obtained by testing for independence of the error variables. This is a fairly important test that checks if the LiNGAM assumption holds true for the proposed case. Essentially, we want the test to fail at rejecting the null hypothesis, which in this case is the independence of the error/exogenous variables of LiNGAM.
\end{itemize}

\begin{figure}[h!]
    \centering
    \includegraphics[scale = 0.35]{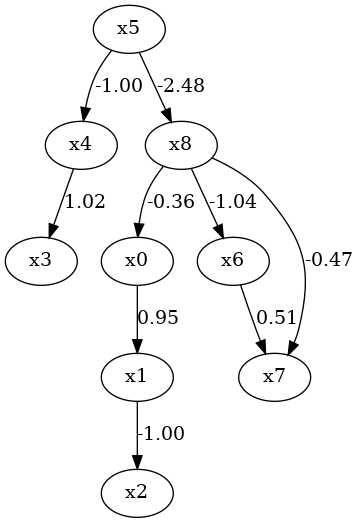}
    \caption{DAG with $w_r = 237$ for FB15K-237 dataset.}
    \label{dag_figure}
\end{figure}
In Fig. \ref{dag_figure}, we see that when $n_{dim} = 3$, the total number of nodes is nine. This is because matrix $\textbf{Q} \in \mathbb{R}^{n_r \times (d_e \times d_e)}$ has $d_e \times d_e$ as columns. So for any dimension $d_e$, there is a quadratic increase in the number of features. This is precisely why we limited ourselves to small values of $d_e$ like 5, 7, and 10.

We further check the validity of the LiNGAM assumption by checking the $p$-values of independence between the error variables of the data matrix. This is because the most important assumption in LiNGAM is that the errors/exogenous variables follow a non-Gaussian distribution and are independent of each other so that there are no latent confounder variables. 

For the experiments, we take $\alpha = 0.01$ of $p$-values. The hypotheses for the proposed results are:\\
$\mathcal{H}_0 \leftarrow$ The error variables are independent of each other.\\
$\mathcal{H}_a \leftarrow$ The error variables are not independent.\\
For all cases, we find that the $p$-values are higher than the $\alpha$ value. This means that the test has failed to reject the null hypothesis.

\section{Limitations}
\label{limit}
The theoretical approach we proposed in the work has its share of limitations. We broadly classify the limitations into two categories:
\begin{itemize}
    \item \textbf{Dimensionality: }This work suffers from the curse of dimensionality. We take the tensor shown in Eq. \eqref{q_tensor_eq} and we project it to a smaller dimension \textbf{Q}. Ignoring the loss of information, one of the things we do to achieve this is by increasing the number of features by two-fold. In this case, by making the matrix have the shape $n_r \times (d_e \times d_e)$. This is especially problematic for DirectLiNGAM due to its polynomial time complexity. 
\begin{figure}[h!]
    \centering
    \includegraphics[scale = 0.5]{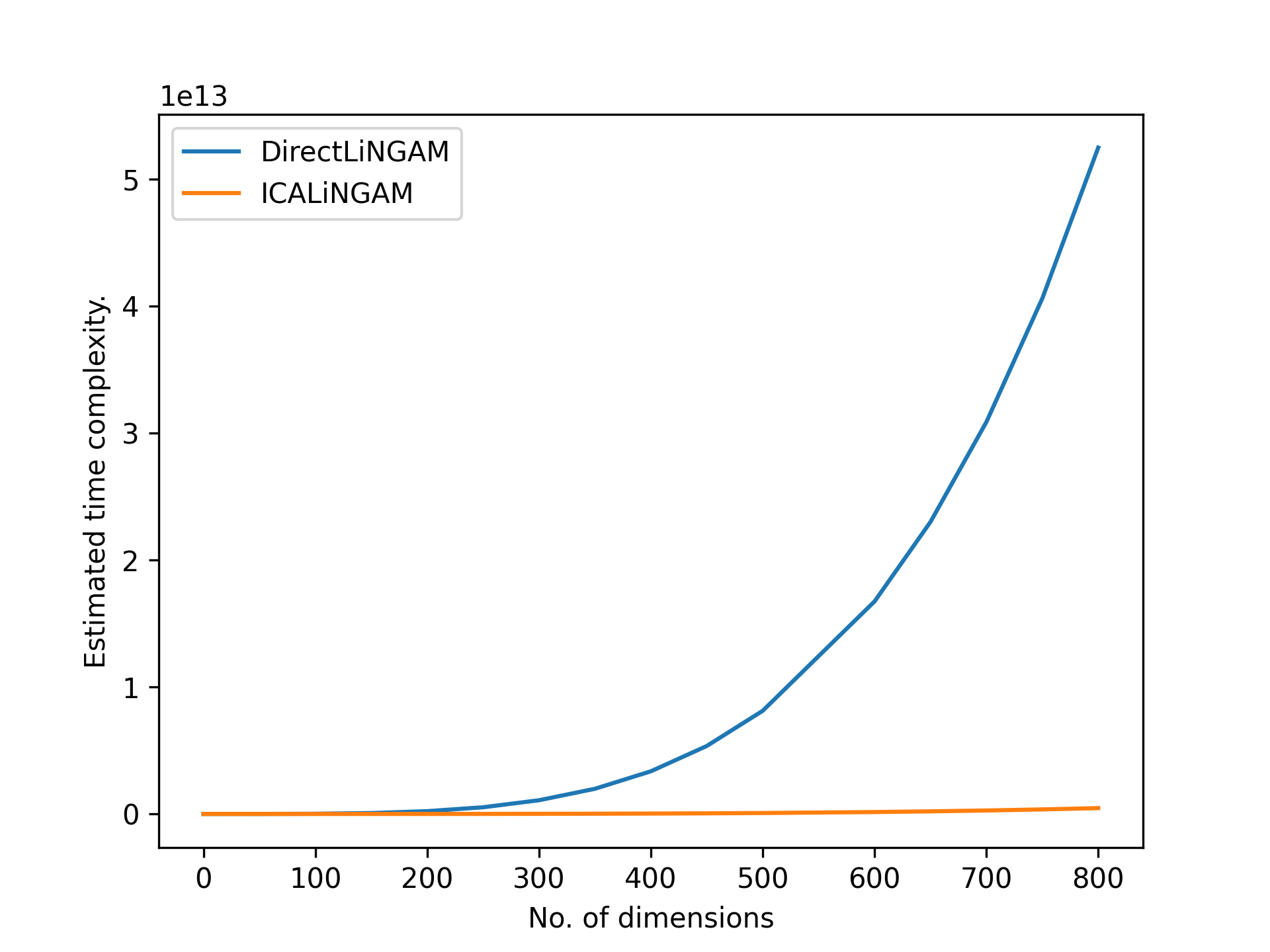}
    \caption{Time complexity  vs. dimensions.}
    \label{fig:time_dim_graph}
\end{figure}

Fig. \ref{fig:time_dim_graph} shows how the increase in the number of dimensions leads to a polynomial increase in the time complexity of the respective algorithm. The complexity of DirectLiNGAM is $\mathcal{O}(np^3M^2 + p^4M^3)$ and that of ICALiNGAM is $\mathcal{O}(np^3 + p^4)$. For plotting the graph, we assume that the values of $n$ and $M$ are constant, where $n$ is the number of samples and $M$ is the maximal rank found by the low-rank decomposition used in the kernel-based independence measure [$M (\ll n)$]. This polynomial increase in the time-complexity of the algorithm with just an increase in the dimension $d_e$ is detrimental in cases where we want to examine large, feature-packed KGs for causality.

Another problem is that standard dimensionality reduction techniques like Principal Component Analysis (PCA) or Singular Value Decomposition (SVD) fail to capture the causal relationships among the variables. 
We address this problem in the future works.
    
    \item \textbf{Extrapolation of useable causal relationships} Proposed method is a definite success when it comes to discovery of causal structures in embedding matrices, but extrapolating from these causal relationships to ensure its application is not something we covered in this work.

In this work, we mostly focus on the methodology to check for causal relationships in a Knowledge Graph.
It is more difficult, considering that we look for relationships among the embedded features of a KG. 
In Fig. \ref{dag_figure}, we see elements $x_0 \rightarrow x_8$. These $9$ variables are $9$ features embedded by KG when we consider dimensionality $d_e \times d_e$. It is not possible to perform an inverse operation to get the required result, as the embeddings for each entity vary greatly. 
Further analysis will give us more data on how data can be embedded in a way where it makes sense to use causal discovery for working with applications like recommendation systems, etc.
We will be able to develop practical applications like causal-enforced reasoning.
\end{itemize}

\section{Conclusions} \label{conclusions}
The approaches presented in this work were the result of us trying to formally explore a previously unexplored avenue of causal discovery. Through the results, we could come to the conclusion that it is, in fact, plausible to discover causal structures in KG. We tried to approach this problem of causal discovery from a purely statistical perspective without the requirement of any background knowledge or problem/dataset specific ontologies. Future improvements include using better methods of tensor decomposition and representation such as LowFER \cite{amin2020lowfer}, which is a generalized form of TuckER, better forms of decomposition and dimensionality reduction that can capture and project the causal structure into lower dimensions.  Though we could get the results for the proposed approach, more in-depth analysis and experimentation is required to bring out the applications for the proposed approach on KG-based tasks (such as link prediction or KG completion).

\bibliographystyle{plainnat}
\bibliography{references}

\begin{thebibliography}{41}
\providecommand{\natexlab}[1]{#1}
\providecommand{\url}[1]{\texttt{#1}}
\expandafter\ifx\csname urlstyle\endcsname\relax
  \providecommand{\doi}[1]{doi: #1}\else
  \providecommand{\doi}{doi: \begingroup \urlstyle{rm}\Url}\fi

\bibitem[Akaho(2006)]{akaho2006kernel}
Shotaro Akaho.
\newblock {A Kernel Method for Canonical Correlation Analysis}.
\newblock \emph{CoRR}, abs/0609071, 2006.

\bibitem[Amin et~al.(2020)Amin, Varanasi, Dunfield, and
  Neumann]{amin2020lowfer}
Saadullah Amin, Stalin Varanasi, Katherine~Ann Dunfield, and G{\"u}nter
  Neumann.
\newblock {LowFER: Low-rank Bilinear Pooling for Link Prediction}.
\newblock In \emph{ICML}, pages 257--268, 2020.

\bibitem[Auer et~al.(2007)Auer, Bizer, Kobilarov, Lehmann, Cyganiak, and
  Ives]{dbpedia}
S\"{o}ren Auer, Christian Bizer, Georgi Kobilarov, Jens Lehmann, Richard
  Cyganiak, and Zachary Ives.
\newblock {DBpedia: A Nucleus for a Web of Open Data}.
\newblock In \emph{ISWC}, page 722–735, 2007.

\bibitem[Bach and Jordan(2002)]{bach2002kernel}
Francis~R Bach and Michael~I Jordan.
\newblock {Kernel Independent Component Analysis}.
\newblock \emph{JMLR}, Jul:\penalty0 1--48, 2002.

\bibitem[Balažević et~al.(2019)Balažević, Allen, and
  Hospedales]{Balazevic2019}
Ivana Balažević, Carl Allen, and Timothy~M. Hospedales.
\newblock {TuckER: Tensor Factorization for Knowledge Graph Completion}.
\newblock In \emph{EMNLP-IJCNLP}, pages 5185--5194, November 2019.

\bibitem[Bollen and Paxton(1998)]{bollen1998interactions}
Kenneth~A Bollen and Pamela Paxton.
\newblock {Interactions of Latent Variables in Structural Equation Models}.
\newblock \emph{Structural Equation Modeling: A Multidisciplinary Journal},
  5\penalty0 (3):\penalty0 267--293, 1998.

\bibitem[Bordes et~al.(2013)Bordes, Usunier, Garcia-Duran, Weston, and
  Yakhnenko]{fb15k}
Antoine Bordes, Nicolas Usunier, Alberto Garcia-Duran, Jason Weston, and Oksana
  Yakhnenko.
\newblock {Translating Embeddings for Modeling Multi-relational Data}.
\newblock In \emph{NIPS}, pages 1--9, 2013.

\bibitem[Dettmers et~al.(2018)Dettmers, Minervini, Stenetorp, and
  Riedel]{wn18rr}
Tim Dettmers, Pasquale Minervini, Pontus Stenetorp, and Sebastian Riedel.
\newblock {Convolutional 2D Knowledge Graph Embeddings}.
\newblock In \emph{AAAI}, volume~32, 2018.

\bibitem[Dodge and Rousson(2001)]{Dodge2001}
Yadolah Dodge and Valentin Rousson.
\newblock {On Asymmetric Properties of the Correlation Coeffcient in the
  Regression Setting}.
\newblock \emph{The American Statistician}, 55\penalty0 (1):\penalty0 51--54,
  Feb 2001.

\bibitem[Dodge and Yadegari(2009)]{Dodge2009}
Yadolah Dodge and Iraj Yadegari.
\newblock {On Direction of Dependence}.
\newblock \emph{Metrika}, 72\penalty0 (1):\penalty0 139--150, Aug 2009.

\bibitem[Harris et~al.(2020)Harris, Millman, van~der Walt, Gommers, Virtanen,
  Cournapeau, Wieser, Taylor, Berg, Smith, Kern, Picus, Hoyer, van Kerkwijk,
  Brett, Haldane, del R{\'{i}}o, Wiebe, Peterson, G{\'{e}}rard-Marchant,
  Sheppard, Reddy, Weckesser, Abbasi, Gohlke, and Oliphant]{harris2020array}
Charles~R. Harris, K.~Jarrod Millman, St{\'{e}}fan~J. van~der Walt, Ralf
  Gommers, Pauli Virtanen, David Cournapeau, Eric Wieser, Julian Taylor,
  Sebastian Berg, Nathaniel~J. Smith, Robert Kern, Matti Picus, Stephan Hoyer,
  Marten~H. van Kerkwijk, Matthew Brett, Allan Haldane, Jaime~Fern{\'{a}}ndez
  del R{\'{i}}o, Mark Wiebe, Pearu Peterson, Pierre G{\'{e}}rard-Marchant,
  Kevin Sheppard, Tyler Reddy, Warren Weckesser, Hameer Abbasi, Christoph
  Gohlke, and Travis~E. Oliphant.
\newblock {Array Programming with NumPy}.
\newblock \emph{Nature}, 585\penalty0 (7825):\penalty0 357--362, Sep 2020.

\bibitem[Hofer-Szab{\'o} et~al.(1999)Hofer-Szab{\'o}, R{\'e}dei, and
  Szab{\'o}]{hofer1999reichenbach}
G{\'a}bor Hofer-Szab{\'o}, Mikl{\'o}s R{\'e}dei, and L{\'a}szl{\'o}~E
  Szab{\'o}.
\newblock {On Reichenbach's Common Cause Principle and Reichenbach's Notion of
  Common Cause}.
\newblock \emph{The British Journal for the Philosophy of Science}, 50\penalty0
  (3):\penalty0 377--399, 1999.

\bibitem[Hyvärinen and Oja(2000)]{Hyvaerinen2000}
A.~Hyvärinen and E.~Oja.
\newblock {Independent Component Analysis: Algorithms and Applications}.
\newblock \emph{Neural Networks}, 13\penalty0 (4-5):\penalty0 411--430, Jun
  2000.

\bibitem[Jenatton et~al.(2012)Jenatton, Roux, Bordes, and
  Obozinski]{NIPS2012_0a1bf96b}
Rodolphe Jenatton, Nicolas Roux, Antoine Bordes, and Guillaume~R Obozinski.
\newblock {A Latent Factor Model for Highly Multi-relational Data}.
\newblock In \emph{NIPS}, 2012.

\bibitem[Ji et~al.(2021)Ji, Pan, Cambria, Marttinen, and Yu]{Ji2021tnnls}
Shaoxiong Ji, Shirui Pan, Erik Cambria, Pekka Marttinen, and Philip~S. Yu.
\newblock {A Survey on Knowledge Graphs: Representation, Acquisition and
  Applications}.
\newblock \emph{IEEE TNNLS}, pages 1--21, 2021.

\bibitem[Keith et~al.(2020)Keith, Jensen, and O{'}Connor]{keith-etal-2020-text}
Katherine Keith, David Jensen, and Brendan O{'}Connor.
\newblock {Text and Causal Inference: A Review of Using Text to Remove
  Confounding from Causal Estimates}.
\newblock In \emph{ACL}, July 2020.

\bibitem[Lin et~al.(2015)Lin, Liu, Sun, Liu, and Zhu]{TransR}
Yankai Lin, Zhiyuan Liu, Maosong Sun, Yang Liu, and Xuan Zhu.
\newblock {Learning Entity and Relation Embeddings for Knowledge Graph
  Completion}.
\newblock In \emph{AAAI}, page 2181–2187, 2015.

\bibitem[Liu et~al.(2016)Liu, Jiang, Evdokimov, Ling, Zhu, Wei, and
  Hu]{liu2016probabilistic}
Quan Liu, Hui Jiang, Andrew Evdokimov, Zhen-Hua Ling, Xiaodan Zhu, Si~Wei, and
  Yu~Hu.
\newblock {Probabilistic Reasoning via Deep Learning: Neural Association
  Models}.
\newblock \emph{CoRR}, abs/1603.07704, 2016.

\bibitem[Minker(1982)]{minker1982indefinite}
Jack Minker.
\newblock {On Indefinite Databases and the Closed World Assumption}.
\newblock In \emph{International Conference on Automated Deduction}, pages
  292--308. Springer, 1982.

\bibitem[Munch et~al.(2019)Munch, Dibie, Wuillemin, and
  Manfredotti]{semex_2019_4}
Melanie Munch, Juliette Dibie, Pierre-Henri Wuillemin, and Cristina
  Manfredotti.
\newblock {Interactive Causal Discovery in Knowledge Graphs}.
\newblock In \emph{PROFILES-SEMEX}, pages 78--93, 2019.

\bibitem[Nickel et~al.(2011)Nickel, Tresp, and Kriegel]{rescal}
Maximilian Nickel, Volker Tresp, and Hans-Peter Kriegel.
\newblock {A Three-Way Model for Collective Learning on Multi-Relational Data}.
\newblock In \emph{ICML}, 2011.

\bibitem[Nickel et~al.(2012)Nickel, Tresp, and Kriegel]{yago_factorizing}
Maximilian Nickel, Volker Tresp, and Hans-Peter Kriegel.
\newblock {Factorizing YAGO: Scalable Machine Learning for Linked Data}.
\newblock In \emph{WWW}, page 271–280, 2012.

\bibitem[Nickel et~al.(2016)Nickel, Rosasco, and Poggio]{HolE}
Maximilian Nickel, Lorenzo Rosasco, and Tomaso Poggio.
\newblock {Holographic Embeddings of Knowledge Graphs}.
\newblock In \emph{AAAI}, page 1955–1961, 2016.

\bibitem[Pang et~al.(2015)Pang, Wang, Cheng, Zhang, and
  Li]{pang2015topological}
Chaoyi Pang, Junhu Wang, Yu~Cheng, Haolan Zhang, and Tongliang Li.
\newblock {Topological Sorts on DAGs}.
\newblock \emph{Information Processing Letters}, 115\penalty0 (2):\penalty0
  298--301, 2015.

\bibitem[Pearl(2009)]{Pearl2009}
Judea Pearl.
\newblock \emph{Causality}.
\newblock Cambridge University Press, 2009.

\bibitem[Schaechtle et~al.(2013)Schaechtle, Stathis, and
  Bromuri]{schaechtle2013multi}
Ulrich Schaechtle, Kostas Stathis, and Stefano Bromuri.
\newblock {Multi-dimensional Causal Discovery}.
\newblock In \emph{IJCAI}, 2013.

\bibitem[Schwarz et~al.(1978)]{schwarz1978estimating}
Gideon Schwarz et~al.
\newblock {Estimating the Dimension of a Model}.
\newblock \emph{Annals of Statistics}, 6\penalty0 (2):\penalty0 461--464, 1978.

\bibitem[Shimizu et~al.(2006)Shimizu, Hoyer, Hyvarinen, and
  Kerminen]{JMLR:v7:shimizu06a}
Shohei Shimizu, Patrik~O. Hoyer, Aapo Hyvarinen, and Antti Kerminen.
\newblock {A Linear Non-Gaussian Acyclic Model for Causal Discovery}.
\newblock \emph{JMLR}, 7\penalty0 (72):\penalty0 2003--2030, 2006.

\bibitem[Shimizu et~al.(2011)Shimizu, Inazumi, Sogawa, Hyvarinen, Kawahara,
  Washio, Hoyer, and Bollen]{Shimizu2011}
Shohei Shimizu, Takanori Inazumi, Yasuhiro Sogawa, Aapo Hyvarinen, Yoshinobu
  Kawahara, Takashi Washio, Patrik~O. Hoyer, and Kenneth Bollen.
\newblock {DirectLiNGAM: A Direct Method for Learning A Linear Non-Gaussian
  Structural Equation Model}.
\newblock \emph{JMLR}, 12:\penalty0 1225–1248, July 2011.

\bibitem[Socher et~al.(2013)Socher, Chen, Manning, and Ng]{NTN}
Richard Socher, Danqi Chen, Christopher~D. Manning, and Andrew~Y. Ng.
\newblock {Reasoning with Neural Tensor Networks for Knowledge Base
  Completion}.
\newblock In \emph{NIPS}, 2013.

\bibitem[Spirtes et~al.(2000)Spirtes, Glymour, Scheines, and
  Heckerman]{spirtes2000causation}
Peter Spirtes, Clark~N Glymour, Richard Scheines, and David Heckerman.
\newblock \emph{{Causation, Prediction, and Search}}.
\newblock MIT press, 2000.

\bibitem[Suchanek et~al.(2007)Suchanek, Kasneci, and Weikum]{yago}
Fabian~M Suchanek, Gjergji Kasneci, and Gerhard Weikum.
\newblock {YAGO: A Core of Semantic Knowledge}.
\newblock In \emph{WWW}, pages 697--706, 2007.

\bibitem[Toutanova et~al.(2015)Toutanova, Chen, Pantel, Poon, Choudhury, and
  Gamon]{fb15k-237}
Kristina Toutanova, Danqi Chen, Patrick Pantel, Hoifung Poon, Pallavi
  Choudhury, and Michael Gamon.
\newblock {Representing Text for Joint Embedding of Text and Knowledge Bases}.
\newblock In \emph{EMNLP}, pages 1499--1509, September 2015.

\bibitem[Trouillon et~al.(2016)Trouillon, Welbl, Riedel, Gaussier, and
  Bouchard]{ComplEx}
Th\'{e}o Trouillon, Johannes Welbl, Sebastian Riedel, \'{E}ric Gaussier, and
  Guillaume Bouchard.
\newblock {Complex Embeddings for Simple Link Prediction}.
\newblock In \emph{ICML}, page 2071–2080, 2016.

\bibitem[Wang and Song(2017)]{wang2017tensor}
Miaoyan Wang and Yun Song.
\newblock {Tensor Decompositions via Two-mode Higher-order SVD (HOSVD)}.
\newblock In \emph{AISTATS}, pages 614--622. PMLR, 2017.

\bibitem[Wang et~al.(2017)Wang, Mao, Wang, and Guo]{wang2017knowledge}
Quan Wang, Zhendong Mao, Bin Wang, and Li~Guo.
\newblock {Knowledge Graph Embedding: A Survey of Approaches and Applications}.
\newblock \emph{IEEE TKDE}, 29\penalty0 (12):\penalty0 2724--2743, 2017.

\bibitem[Wang et~al.(2014)Wang, Zhang, Feng, and Chen]{wang2014knowledge}
Zhen Wang, Jianwen Zhang, Jianlin Feng, and Zheng Chen.
\newblock {Knowledge Graph Embedding by Translating on Hyperplanes}.
\newblock In \emph{AAAI}, 2014.

\bibitem[Yang et~al.(2015)Yang, Yih, He, Gao, and Deng]{yang2015embedding}
Bishan Yang, Scott Wen-tau Yih, Xiaodong He, Jianfeng Gao, and Li~Deng.
\newblock {Embedding Entities and Relations for Learning and Inference in
  Knowledge Bases}.
\newblock In \emph{ICLR}, May 2015.

\bibitem[Yang et~al.(2019)Yang, Liu, Chen, Ma, and Tsai]{yang2019causal}
Chao-Han~Huck Yang, Yi-Chieh Liu, Pin-Yu Chen, Xiaoli Ma, and Yi-Chang~James
  Tsai.
\newblock {When Causal Intervention meets Adversarial Examples and Image
  Masking for Deep Neural Networks}.
\newblock In \emph{ICIP}, pages 3811--3815, 2019.

\bibitem[Yu et~al.(2021)Yu, Chhetri, Canedo, Goyal, and Faruque]{pykg2vec}
Shih-Yuan Yu, Sujit~Rokka Chhetri, Arquimedes Canedo, Palash Goyal, and
  Mohammad Abdullah~Al Faruque.
\newblock {Pykg2vec: A Python Library for Knowledge Graph Embedding}.
\newblock \emph{JMLR}, 22\penalty0 (16):\penalty0 1--6, 2021.

\bibitem[Zhu and Ngo(2020)]{zhu2020cookgan}
Bin Zhu and Chong-Wah Ngo.
\newblock {CookGAN: Causality based Text-to-Image Synthesis}.
\newblock In \emph{CVPR}, pages 5519--5527, 2020.

\end{thebibliography}


\end{document}